%% file: main.tex
\documentclass{article} 
\usepackage{iclr2025_conference, times}

\input{math_commands.tex}

\usepackage{microtype}
\usepackage{hyperref}
\usepackage{url}
\usepackage{booktabs}
\usepackage{wrapfig}

\title{SelectFormer: Private and Practical Data Selection for Transformers}

\iclrfinalcopy

\author{Xu Ouyang, Felix Xiaozhu Lin, Yangfeng Ji \\
Department of Computer Science\\
University of Virginia\\
\texttt{\{ftp8nr, felixlin, yangfeng\}@virginia.edu}
}

\input{header.tex}
\begin{document}
\maketitle 
\input{abs}
\input{intro}		
\input{motiv}
\input{related}		
\input{detail}		
\input{eval}		
\input{summary}		

\bibliography{iclr2025_conference}
\bibliographystyle{iclr2025_conference}

\input{appendix}
\end{document}

%% file: math_commands.tex
\usepackage{amsmath,amsfonts,bm}

\def\eqref#1{equation~\ref{#1}}

\def\1{\bm{1}}

\DeclareMathAlphabet{\mathsfit}{\encodingdefault}{\sfdefault}{m}{sl}
\SetMathAlphabet{\mathsfit}{bold}{\encodingdefault}{\sfdefault}{bx}{n}



%% file: header.tex
\usepackage{wrapfig}
\usepackage{comment}
\usepackage{colortbl}

\usepackage{hyperref}

\usepackage{amsthm} 

\usepackage{color,soul}
\usepackage{graphics}
\usepackage{hyperref}
\usepackage{lipsum}
\usepackage{tikz}
\usepackage{enumitem}	
\usepackage{listings} 
\usepackage{booktabs}	
\usepackage{lipsum}

\usepackage{capt-of}	
\usepackage{diagbox}
\usepackage{multirow}
\usepackage{todonotes}  

\usepackage{mathrsfs}	
\usepackage{amsfonts}

\usepackage[export]{adjustbox} 

\usepackage{algorithm,amsmath,algpseudocode} 
\usepackage{boldline}					

\usepackage{multirow}

\definecolor{frenzyorange}{RGB}{249, 158, 26}

\renewcommand{\paragraph}[1]{\vskip 3pt\noindent\textbf{#1 }}	 

  {\begin{list}{$\bullet$}%
     {\setlength{\parsep}{0pt}%
      \setlength{\topsep}{0pt}%
      \setlength{\itemsep}{2pt}}}%
  {\end{list}}
\newcommand\Noted[1]{} 

\newcommand\xzlNote[1]{\sethlcolor{yellow} \hl{#1}} 

\definecolor{darkblue}{rgb}{0.0, 0.0, 0.55}
\definecolor{mygreen}{HTML}{ADFF2F}
\definecolor{mylightgray}{gray}{0.8}

\newenvironment{myitemize}%
  {\begin{itemize}
	[leftmargin=0cm,
		itemindent=.3cm,
		labelwidth=\itemindent,
		labelsep=0pt,
		parsep=1pt,
		topsep=1pt,
		itemsep=1pt,
		align=left]
  }%
  {\end{itemize}}    

\newenvironment{myenumerate}%
  {\begin{enumerate}
	[leftmargin=.cm,itemindent=.5cm,labelwidth=\itemindent,
		labelsep=0pt,
		parsep=1pt,
		topsep=1pt,
		itemsep=3pt,
		align=left]
  }%
  {\end{enumerate}}    

\newcommand\sect[1]{Section~\ref{sec:#1}}	

\input{terms}

\makeatletter
\def\@copyrightspace{\relax}
\makeatother

%% file: terms.tex
\newcommand{\sys}{SelectFormer}
\newcommand{\mps}{MPS}
\newcommand{\Mps}{MPS}

\newcommand{\sps}{SPS}

\newcommand{\mf}{MPCF{\scriptsize ORMER}}

\newcommand{\target}{$M_{target}$}

\newcommand{\proxy}{proxy}

\newcommand{\oracle}{\textit{Oracle}}
\newcommand{\Oracle}{\textit{Oracle}}


%% file: abs.tex
\begin{abstract}


Critical to a free data market
is \textit{private data selection}, 
i.e. the model owner selects and then appraises training data from the data owner before both parties commit to a transaction.
To keep the data and model private, 
this process shall evaluate the target model to be trained over Multi-Party Computation (MPC). 
While prior work suggests that evaluating Transformer-based models over MPC is prohibitively expensive, 
this paper makes it practical for the purpose of data selection. 
Our contributions are three:
(1) a new pipeline for private data selection over MPC; 
(2) emulating high-dimensional nonlinear operators with low-dimension MLPs, which are trained on a small sample of the data of interest; 
(3) scheduling MPC in a parallel, multiphase fashion. 
We evaluate our method on diverse Transformer models and NLP/CV benchmarks. 
Compared to directly evaluating the target model over MPC,
our method reduces the delay from thousands of hours to tens of hours, 
while only seeing around 0.20\% accuracy degradation from training with the selected data. 

\end{abstract}

%% file: intro.tex
\section{Introduction}
\label{sec:intro}


\paragraph{Data selection \& appraisal}
In today's ML ecosystem, 
data and models are often owned by separate parties. 
Examples include mobile users versus app providers, 
and small businesses versus marketing firms.
As model owners are interested in acquiring data to train their models from data owners, 
data becomes a commodity to be valued and traded.

\input{fig-selection-flow}
In a free data market, 
no purchase commitment should be required unless both parties reach an agreement, 
for which pre-purchase assessment is vital. 
It contains two steps: select the most valuable data; (optionally) appraise the selected data.
To model owners, pre-purchase selection is key to cost-effectiveness. 
Extensive work already showed that datasets are often redundant and noisy \citep{Settles2012Activelearning,katharopoulos2018not,mirzasoleiman2020coresets, ouyang2022efficient}; 
purchase budgets should be spent on data points that are most likely to maximize model accuracy. 
To data owners, selection allows them to reveal less data in exchange for the same monetary reward, 
which retains their data ownership and reduces privacy exposure. 
The need for judicious data selection is further highlighted by that data candidates are often \textit{unlabeled}~\citep{Hoi2006MedicalImage,Smith2018LessIsMore} and have \textit{skewed label distributions} \citep{kaur2019systematic}. 
This paper targets this issue. 

\paragraph{Challenge: private selection}
Data selection and appraisal should be private: 
they should not leak any data and model parameters; 
if a purchase agreement is eventually reached, the model owner should learn no more beyond the data she is paying for. 
To select and appraise data, 
a common algorithm is to run a model's forward passes over data candidates and select based on the outcome (see \sect{motiv} for details). 
It can be made private by building atop Multi-Party Computation (MPC) \citep{Goldreich1987secretshare,shamir1979share,yao1986generate}:
both parties (owners) jointly evaluate 
over the candidate data, learning only the indices of selected data and the quality measurement. 



However, evaluating Transformer models on MPC is expensive. 
A forward pass of BERT on a batch of 4 requires 3252 communication rounds and over 245 GB data exchanged, taking around 0.76 hours across commodity servers with GPUs. 
Most of the cost comes from executing nonlinear arithmetics such as softmax, 
which is pervasive in Transformers. Some efficient Transformers works can improve inference speed \citep{NEURIPS2021_6ce8d8f3, you2022supertickets, fu2022contrastive}. But to support Transformer inference over MPC, 
recent work approximates nonlinearity with cheaper linear operations \citep{li2022mpcformer,dong2023puma, Chen_2021_CVPR}.
For data selection, 
they are not only too slow (hundreds of GPU hours to select from tens of K data points) but also have poor results as this paper will show.



\paragraph{Goal \& techniques}
Our goal is to accelerate private data selection for transformer-based models, e.g. BERT and ViT, while retaining the training performance. 
Our key insight is that the nonlinearity in Transformers can be fused and evaluated at low dimensions; 
fortunately, the resultant model evaluation outcomes will be acceptable for the purpose of data selection, which only depends on how the outcomes for individual data candidates compare to each other. 
We present a holistic selection pipeline with three key techniques. 




\textit{Nonlinearity emulated by multi-layer perceptron (MLPs)}. 
While existing MPC frameworks treat individual nonlinear operations in isolation (e.g. Newton-Raphson iterations for reciprocal), 
we fuse adjacent non-linear operators and emulate them altogether with 
a small MLP. 
The benefits are twofold.
(1) MLPs significantly reduce MPC costs because they not only serve as approximators (converting nonlinear to linear operations) but also dimension reducers, 
e.g. substituting a 512-dimensional softmax with 2-dimensional MLP.
This sets us apart from prior work that approximates a single nonlinear operation (e.g. reciprocal) with an MLP \citep{chen-etal-2022-x}, 
 which does not reduce dimensions and therefore executes much slower. 
(2) MLPs are data-driven. 
They are trained atop the distributions of the actual model activations, 
tailored to the datasets of interest. 
They only require a small amount of training data, as few as several hundred on
average in our experiment. 


\textit{Multi-phase selection}. 
For efficiency, 
the data/model owners jointly select data in multiple phases.
While early phases evaluate cheap selection models to filter most of the data, 
later phases run more expensive selection models to select from the data that survived from earlier phases. 
This reduces the total selection delay without compromising accuracy. 

\textit{Parallel MPC executions}. 
To further reduce the selection delay, the data/model owners batch network latency-bound MPC operations 
and execute them in parallel to network bandwidth-bound MPC operations; 
they further overlap the MPC communication and computation across different batches.
No prior MPC systems exploit such parallelism, to the best of our knowledge. 

\paragraph{Workflow} There are three stages in our proposed workflow: two in the clear and one over MPC.
As shown in \autoref{fig:selection-flow}, in the beginning, the model owner and the data owner exchange some basic information in the clear. For instance, the total amount of data and the amount of data that the model owner plans to purchase. 
Next is the private multi-phase selection which is over MPC. Two parties secretly share encrypted proxy model parameters and data to do a forward pass and get the entropy value. We then rank them and select the top candidates based on their rank. The remaining data points' indices will be reused in the next phase. The comparison outcomes will be revealed and the data indices are in the clear. Data will not be transferred among phases.
The final transaction is in the clear. After private data selection, the model owner pays for the data it wants and the data owner sends out the data to seal the deal. More details in \sect{detail}.

\paragraph{Results}
We test \sys{} with four target models, seven NLP/CV benchmarks, and a variety of purchase budgets (20\%--40\%).
We show that with commodity GPUs and typical Internet, 
our pipeline can select from 10K--100K data points within tens of hours, one order of magnitude faster than prior work. 
Using unmodified MPC protocol and framework, we provide a strong privacy guarantee. 
Our selected data only results in just 0.20\% lower accuracy compared to 
directly evaluating the target model over MPC for selection, 
a gold method which is however 204$\times $slower. 

\paragraph{Contributions} 
This work proposes a novel application of MPC -- private data selection, and makes it practical on large Transformer models. 
It achieves so via three new techniques: 
\begin{myitemize}
\item  A data-driven MLP approximation that is uniquely tailored for data selection, instead of generic inference. 

\item Multi-phase selection that progressively increases the selection models' capability and reduces the amount of data candidates. 

\item Parallel MPC executions that hide the delays of computation and communication rounds behind that of communication data exchange. 


\end{myitemize}





%% file: fig-selection-flow.tex

\begin{figure}[ht]
	\centering
	\includegraphics[width=0.6\textwidth]{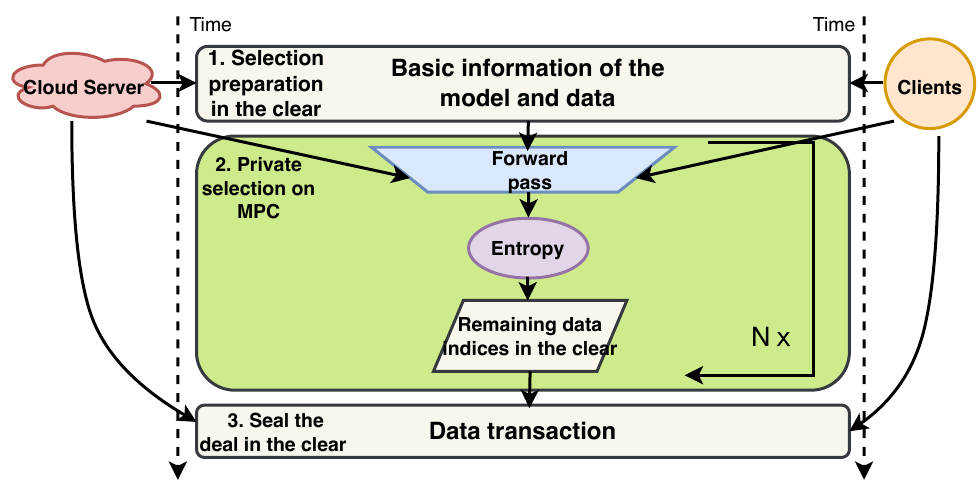}
	\caption{Three stages of our data selection workflow in chronological order.}
	\label{fig:selection-flow}
\end{figure}

%% file: motiv.tex

\section{Background}
\label{sec:motiv}

\subsection{Assumptions on systems}
\paragraph{Two parties}
The model owner possesses a pretrained Transformer model \target{} and wants to finetune it. 
The data owner owns the dataset $D$ for selection. 
The model owner plans to buy $B$ datapoints ($B<|D|$) from dataset $D$ which will be used to finetune the target model $M_{target}$. 
The model owner has a private, small validation set, on which she wants to maximize the validation accuracy. 
The two parties agree on $B$ and that they will use MPC for selection. 




\paragraph{Datapoints}
We follow a common premise in prior work on data selection: \citep{coleman2019selection,Mahmood2022OptimizingDataCollection}, 
the dataset $D$ is \textbf{unlabeled} and might have an \textbf{imbalanced} class distribution, which warrants careful selection before training. 
This premise is motivated by real-world situations where data labels are scarce
and data owners often lack the motivation, expertise, and authority to label their data \citep{Hoi2006MedicalImage,Smith2018LessIsMore}. 

\paragraph{Threat model}
Without alternation, we follow a semi-honest model widely used in prior security research \citep{knott2021crypten,dong2023puma,mohassel2017secureml}: 
both parties faithfully follow the MPC 
protocol but nevertheless attempt to learn each other's private information (model weights and datapoints) through their interactions. 
Notably, we do not deviate from protocols implemented by well-known frameworks such as Crypten \citep{knott2021crypten}. 
As such, our privacy guarantee is as strong as the underlying framework. 

\subsection{Algorithm foundations}
\input{fig-transformer-analysis}
\textbf{(1) Maximum entropy selection} 
Active learning (AL) \citep{Settles2012Activelearning} 
selects from unlabeled data to maximize training performance. 
To appraise an example $x$ for training model $M$, 
AL runs a forward pass of $M$ over $x$ (or query $x$ with $M$) and computes the prediction entropy: 
higher entropy implies lower model confidence, hence a higher learning value of $x$. 
\textbf{(2) Proxy models} 
Querying all examples in $D$ with $M_{target}$ can be slow. 
To this end, prior work uses a lightweight model $M_p$ called ``proxy'' for selection before training $M_{target}$ with the selected data~\citep{coleman2019selection}. 
To select valuable data, $M_p$'s prediction should resemble that of $M_{target}$; 
hence $M_p$ is often created as a miniature version of $M_{target}$: 
examples include ResNet-18 as proxy for ResNet-50 \citep{coleman2019selection}. 
We use simplified Transformers as proxy models with fewer layers and attention heads.

\subsection{Transformers, MPC, and overhead analysis}



\paragraph{Transformers} 

We focus on transformer-based models \citep{Ashish2017transformer}. Each layer of these models features a Multihead Attention paired with FeedForward. Both are succeeded by layer normalization and a residual connection. Within the Multihead Attention, operations proceed sequentially from QKV linear operation to the attention mechanism, and finally to an attention output.


\paragraph{MPC setting: 2PC} 
We assume the most common MPC setting: secret sharing under two-party computation (2PC) \citep{Qiang20192PC}. 
The model owner randomly decomposes $x$ into two shares, $x_1$ and $x_2$ (such that $x_1 + x_2 = x$), retains $x_1$, and convey $x_2$ to the data owner. Analogously, the data owner fragments $y$ into $y_1$ and $y_2$ and transfers $y_2$ to the model owner. These shares individually disclose no information about the original numbers, $x$ and $y$. Reconstruction of a number is possible when both parties exchange their shares and sum them up.

To compute $z = x + y$, each party add their shares of $x$ and $y$, yielding $z_1 = x_1 + y_1$ and $z_2 = x_2 + y_2$. Both parties can retrieve $z$ by adding $z_1$ and $z_2$. For multiplication, The parties can offline generate a random triple called Beaver triples \citep{Beavor1992}, $a$, $b$, and $c$ (such that $ab = c$). The elements of this triple are then partitioned and disseminated amongst the parties, akin to $x$ and $y$. The model owner calculates $\epsilon_1 = x_1 - a_1 $, $\delta_1 = y_1 - b_1$ while the data owner computes $\epsilon_2 = x_2 - a_2 $, $\delta_2 = y_2 - b_2$. The parties jointly reconstruct $\epsilon$ and $\delta$. Subsequently, the model owner computes $p_1 = c_1 + \epsilon y_1 + \delta x_1 + \epsilon \delta$ and the data owner calculates $p_2 = c_2 + \epsilon y_2 + \delta x_2$, enabling the retrieval of $p = xy$. Throughout this procedure, all intermediate computations, including $\epsilon_1$, $\epsilon_2$, $\delta_1$ and $\delta_2$ maintain the confidentiality of $x$, $y$ and $p$, precluding any information leakage.
\paragraph{Major overheads} 
comes from that MPC lacks native support for nonlinear operations, such as softmax, logarithmic, and exponential, which are pervasive in Transformers.
In response, MPC frameworks provide corresponding linear approximations. 
Unfortunately, the overhead is still high. 
\autoref{fig:transformer-analysis} shows the cost of the operations for \textit{one} transformer block. 
As shown in the table, the cost of nonlinear arithmetic dominates. 
Notably, softmax contributes 81.9\% communication data and 142 communication rounds. 
The overheads become prohibitive for data selection, which needs to evaluate Transformers on tens of thousands of data points over MPC. We show this in \sect{eval}.

%% file: fig-transformer-analysis.tex


\begin{wrapfigure}{r}{0.5\textwidth}
\centering
\includegraphics[width=0.4\textwidth{}]{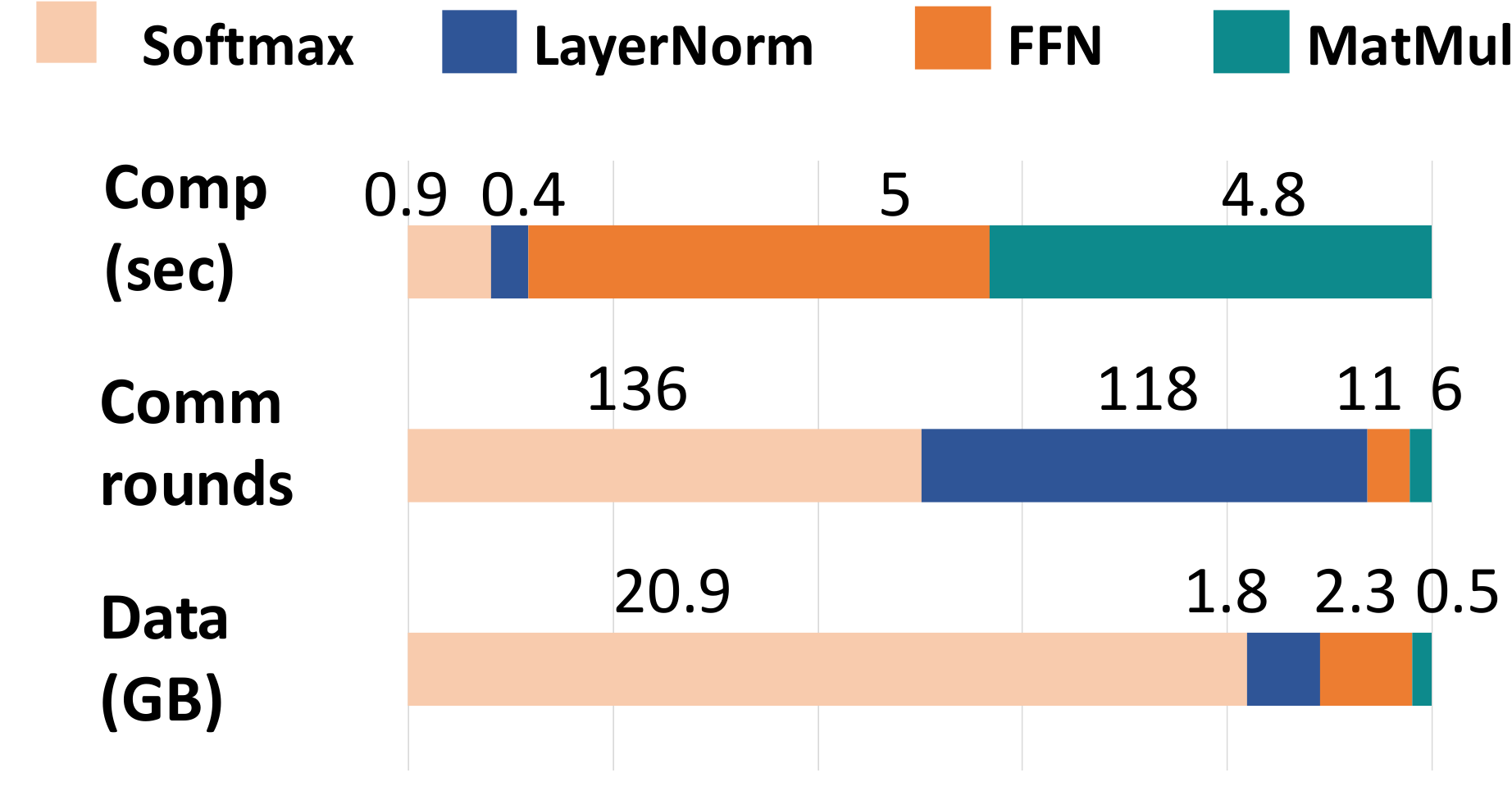}
\vspace{-1em}
\caption{Transformers over MPC incurs high communication and computation overhead. Showing one forward pass of one layer (12 heads) over a batch of 5 (maximum allowable on our GPU). Hardware: Quadro RTX 4000. MPC framework: Crypten \citep{knott2021crypten}.}
\label{fig:transformer-analysis}
\end{wrapfigure} 

%% file: related.tex
\section{Related works}


\paragraph{Data selection and appraisal}
Our work builds on the idea of using a miniature version (proxy) of the target model as the selector~\citep{coleman2019selection}.
Yet, proxies alone are still too expensive for MPC as we will show in \sect{eval}. 
\cite{xu2022data} investigates data appraisal over MPC. 
Contrasting to us, they assume labeled data (a strong assumption in practice), which allows them to appraise data via forward influence functions.   
Unlike us focusing on Transformer models, they only demonstrated logistic regression tasks. 
\cite{mindermann2022prioritized} introduces RHO loss that quantifies by how much data candidates would reduce the loss on a small holdout set if the model were to train the candidates. 
They then select data based on RHO loss. 
Their methods inspire our bootstrap dataset. 
However, they rely on labeled data, whereas we focus on a more challenging setting -- selecting unlabeled data. 

\paragraph{MPC inference of neural networks}
Existing research has developed discretized DNN training with a customized two-party protocol \citep{agrawal2019quotient} and MPC-friendly approximations to speed up CNN. \cite{chou2018faster} develop an optimization framework that minimizes the approximation error of order two polynomial to ReLU. 
\cite{244032} alleviates this problem by using a set of heuristics along with Neural Architecture Search (NAS). 
\cite{mohassel2017secureml} proposes a costly approximation of softmax by replacing exponential with ReLU functions on high-dimension inputs.

\mf~\citep{li2022mpcformer} and PUMA \citep{dong2023puma} directly address efficient Transformer inference over MPC. 
\mf is implemented based on Crypten \citep{knott2021crypten} framework and its key idea is that approximate nonlinear operations in the student model and then distill it; 
PUMA proposes new approximations for faster secure inference in a three-party setting. Bolt \citep{pang2024bolt} proposes a new Transformer inference framework.
Compared to us, they miss the key opportunities of reducing nonlinearity dimensions and using learned approximators (MLPs). 
Furthermore, \mf's model distillation approach is ill-suited to data selection, where labeled data for distillation is lacking. 
As \sect{eval} will show, they perform poorly in data selection. 

%% file: detail.tex
\section{Method}
\label{sec:detail}

\subsection{High-level workflow}

\paragraph{Pre-selection bootstrap:}
Prior to the selection, the two parties exchange meta-information \textit{in the clear}: 
the purchase budget $B$, 
the secure computation framework, the target model architecture type, pre-processing methods,  and the metadata about the $S$, which includes the number of data points.

The model owner purchases bootstrap data $S_{boot}$ which 
constitutes a small fraction of the purchase budget $B$. 
The data owner randomly samples $S_{boot}$ from $S$; 
the purchase requires no selection and incurs no MPC overhead. 
The model owner will use $S_{boot}$ to generate \proxy{} models $\hat{M}_{1..N}$, 
as will be described below. 
\paragraph{Multipass selection}
For efficiency, the selection process is a progressive sieve: 
earlier phases run smaller selector models for quickly filtering a majority of redundancy, 
whereas later phases run larger selector models for slower, more precise selection. 
Concretely, 
both parties jointly execute $N$ selection phases over MPC. 
A phase $i$ downselects dataset $S_{i-1}$ to the set $S_i$ 
with a selectivity $\alpha_i = |S_i| / |S_{i-1}|, \alpha_i <1$. 

In phase $i$, the model owner queries the dataset $S_{i-1}$ 
with a \proxy{} model $\hat{M}_i$.
The query is executed as a forward pass of $\hat{M}_i$ on $S_{i-1}$ on MPC. 
The forward pass computes the prediction entropy values, which are encrypted.
Next, both parties jointly find the \textit{indices} of $|S_i|$ highest entropy values: 
execute the QuickSelect algorithm with $O(|S_{i-1}|)$ pairwise comparisons between entropy values; 
the comparison is over MPC, each taking 8 communication rounds and transferring 432 bytes. 
A comparison does not reveal its inputs (entropy values) but only the binary outcome. 
Finally, the indices of data points with the highest prediction entropy constitute the input $S_i$ for the next phase. 
\input{fig-workflow}
$N$  phases result in the final dataset $S_{N}$. 
The model owner may offer to buy $S_{N}$, or request an appraisal of $S_{N}$ before buying.  
For appraisal, both parties jointly compute an average entropy over $S_{N}$ and reveal the average entropy. 
If the average entropy is sensitive, both parties can jointly compute if the (encrypted) average exceeds a pre-defined threshold and only reveals the one-bit outcome. 
\paragraph{Privacy guarantees}
The following information is kept private: dataset $S$; parameters of \target{} and $\hat{M}_{1..N}$;  
the prediction outcome and entropy.  
The following is revealed: 
the \textit{rank} of prediction entropy; 
the architecture (operations) of $\hat{M}_{1..N}$; 
the number of data points and budget.
\subsection{Generating \proxy{} models in detail} 
\label{sec:design:model}

A \proxy{} model $\hat{M}_i$ is characterized by $<l_i,w_i,d_i>$:
$L_i$ transformer layers with  $A_i$ attention heads each (width), 
in which the nonlinear modules are substituted by MLPs with hidden dimension $d_i$ and the GeLU functions are
substituted by ReLU functions. FFN is also removed from proxy models.
As shown in Figure~\ref{fig:workflow},
the model owner generates \proxy{} models $\hat{M}_{1..N}$ as follows:

\begin{myitemize}
\item Extract a sub-model $M_g$ from $M_{target}$.
$M_g$ comprises $L$ bottom layers from $M_{target}$ where $L=max({l_{1..N}})$,
with all the weights copied over. 
$M_g$ serves as the backbone for $\hat{M}_{1..N}$. 

\item Finetune $M_{g}$ on the bootstrap data $S_{boot}$. 
This serves two purposes: 
(1) collecting sample input/output of $M_{g}$'s nonlinear modules, 
which will be used to train the MLPs that substitute these nonlinear modules; 
(2) roughly adapting $M_{g}$ to a sample from the dataset $S$. 
Initialize $\hat{M}_{1..N}$ by pruning the width and depth of $M_g$. 

\item \textit{Ex vivo} MLP training. 
For each transformer block in a \proxy{} model, 
randomly initialize three MLPs. 
Train them separately on the input/output of nonlinear modules collected from the previous step. More training data following the same distribution as the input/output will be randomly generated as data augmentations. 
See \autoref{sec:design:mlp} for details.

\item \textit{In vivo} MLP training.
Substitute the nonlinear modules in $\hat{M}_i$ with the MLPs trained in the previous step. 
Further finetune the resultant model $\hat{M}_i$ on $S_{boot}$ end-to-end. 
This co-tunes MLPs (the approximate portion of $\hat{M}_i$) and the remaining exact portion. 
\end{myitemize}


The model owner schedules the selection
by setting $\{ <l_i,w_i,d_i> \}_{i=1..N}$ for $N$ phases. 
As described above, the principle of schedule is progressive: increasingly higher $l/w/d$ for later phases. 
Given a budget $B$, \sys{} determines the schedule via offline grid search. See \sect{eval}. 

\input{fig-proxyinside}
\input{tab-e2e}
\subsection{Approximation MLPs for Non-linear Operations}
\label{sec:design:mlp}

Our theoretical foundation is
\cite{hornik1989multilayer}: MLPs are able to approximate any function at any desired degree of accuracy, 
as long as the function is continuous on a closed and bounded subset of $R^n$. 
A transformer's nonlinear modules satisfy this condition \citep{Goodfellow-et-al-2016}. 
We use standard MLPs, each comprising one ReLU layer sandwiched between two linear layers. 
Of a \proxy{} model, MLPs substitute the following nonlinear modules as shown in \autoref{fig:proxyinside}:


\begin{myitemize}
\item \textit{Softmax in the attention module}. 
At each transformer layer, 
a substitute MLP maps along the last dimension of the input 
and
has the same input/output shape as the original softmax operation.

\item \textit{LayerNorm in the attention module}. 
At each transformer layer, 
a substitute MLP emulates the reciprocal operation in LayerNorm. 
Of the LayerNorm, its learnable weights and bias in the affine functions are loaded from the original LayerNorm layers in $M_g$. Its numerator is computed directly on MPC because the total number is a constant and it needs just cheap summation and multiplication.

\item \textit{Softmax over logits and the subsequent entropy computation}.
At the top of the model,
a substitute MLP emulates the two modules combined. 
The MLP input shape is the same as that of softmax. The output will be the entropy itself.

\end{myitemize}

A \proxy{} model of layer $l$ comprises $2l+1$ MLPs in total; 
they have the identical hidden dimension albeit different weights. 
Different \proxy{} models may have MLPs of different hidden dimensions, 
set as part of the schedule as described above. 

\paragraph{Training MLPs}
MLPs need to be trained on the sample input/output of the nonlinear modules being substituted. 
While we can use such input/output observed in finetuning $M_{g}$ over bootstrap data ($S_{boot}$), 
the data amount is inadequate: randomly sampled $S_{boot}$ must be small (3521 datapoints on average in our experiments)
in order to save the budget for private selection. 

We address the problem with the empirical observation that inputs to a nonlinear module largely follow a parametric Gaussian distribution \citep{chen-etal-2022-x}. 
As such, from the actually observed inputs, 
the model owner estimates the Gaussian distribution parameters $\langle \mu,\sigma \rangle$
and uses such parameters to synthesize training inputs and outputs. 

For a given nonlinear module in $M_{g}$, an instance of $\langle \mu,\sigma \rangle$ is estimated and one dataset (5.12 million data points) is synthesized once, 
which is used to train the MLPs that substitute this particular nonlinear module in all \proxy{} models $\hat{M}_{1..N}$. There are three synthesized datasets: $S_{sm}$, $S_{ln}$, and $S_{se}$.
This is shown in Figure~\ref{fig:proxyinside}. 
Note that $\hat{M}_{1..N}$ with MLPs inserted will be further finetuned as described above. The MLP training is very cheap compared with the selection over MPC, which can be measured in minutes.

\subsection{IO scheduling}
Our proxy models perform a sequence of matrix multiplication and ReLU. 
Operating on high-dimensional inputs are bound by the network bandwidth; 
after being projected into lower dimensions,
the operations succeeding the activation functions are bound by network latency. 
To this end, (1) \sys{} stacks and coalesces the latency-bound modules from multiple batches, 
reducing the number of communication rounds and improving the computation throughput.
(2) Furthermore, \sys{} exploits computation and communication parallelism across batches, 
executing a module whenever its needed resources become available. 
For instance, while data exchange is occurring for one batch, computations for the subsequent batches can be performed concurrently. 
Such co-execution is only limited by data dependencies and the available memory of a party to hold operation inputs.

%% file: fig-workflow.tex


\begin{figure}[ht]
\centering
    \includegraphics[width=0.6\textwidth{}]{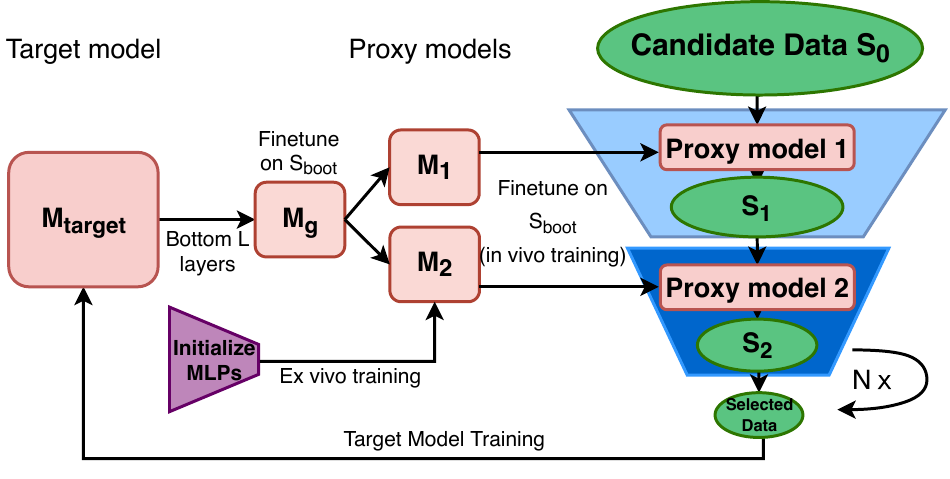}
    \caption{Our high-level workflow. Left: proxy model generation. Right: multi-phase selection. M$_g$ is the base model for the proxy models $\hat{M}_{1..N}$. More details in \autoref{sec:design:model}. }
    \label{fig:workflow}
\end{figure}

%% file: fig-proxyinside.tex
\begin{figure}[ht]
	\centering
	\includegraphics[width=0.7\textwidth]{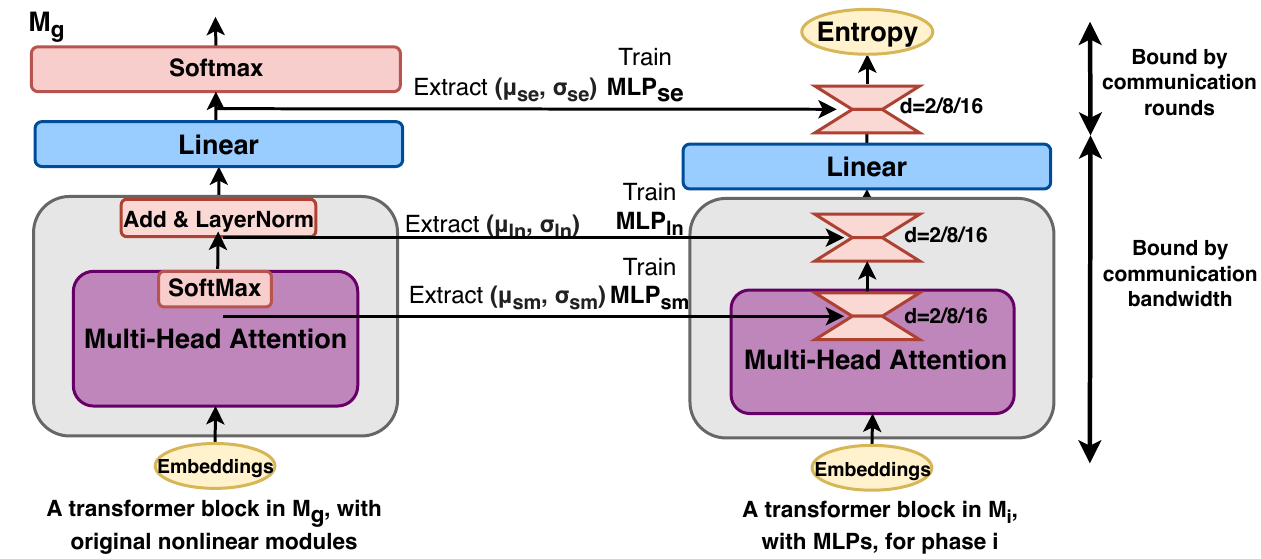}
	\caption{Training MLPs for substituting the non-linearity in Transformer models}
	\label{fig:proxyinside}
	\vspace{-1em}
\end{figure}


%% file: tab-e2e.tex

\begin{table}[]
\centering
\resizebox{\textwidth}{!}{%
\begin{tabular}{@{}l|ccccc|ccccc|cc|cc@{}}
\toprule
                   & \multicolumn{5}{c|}{\textbf{DistilBERT}}                                                                                                                 & \multicolumn{5}{c|}{\textbf{BERT}}                                                                                                                       & \multicolumn{2}{c|}{\textbf{ViT-small}}                      & \multicolumn{2}{c}{\textbf{ViT-base}}                       \\ \cmidrule(l){2-15} 
\multirow{-2}{*}{} & \textbf{SST2}                & \textbf{QNLI}                & \textbf{QQP}                 & \textbf{AGNEWS}              & \textbf{YELP}                & \textbf{SST2}                & \textbf{QNLI}                & \textbf{QQP}                 & \textbf{AGNEWS}              & \textbf{YELP}                & \textbf{Cifar10}             & \textbf{Cifar100}             & \textbf{Cifar10}             & \textbf{Cifar100}            \\ \midrule
Ours               & 82.34                        & 71.46                        & 71.53                        & 88.49                        & 60.18                        & 84.59                        & 73.68                        & 72.73                        & 89.04                        & 61.33                        & 96.46                        & 69.11                         & 96.61                        & 68.42                        \\ \midrule
Random             & 79.08                        & 70.34                        & 66.18                        & 86.43                        & 58.09                        & 79.58                        & 73.56                        & 68.84                        & 87.85                        & 59.16                        & 94.69                        & 43.99                         & 95.63                        & 63.21                        \\
(vs. Ours)         & {\color[HTML]{C00000} -3.26} & {\color[HTML]{C00000} -1.12} & {\color[HTML]{C00000} -5.35} & {\color[HTML]{C00000} -2.06} & {\color[HTML]{C00000} -2.09} & {\color[HTML]{C00000} -5.01} & {\color[HTML]{C00000} -0.12} & {\color[HTML]{C00000} -3.89} & {\color[HTML]{C00000} -1.19} & {\color[HTML]{C00000} -2.17} & {\color[HTML]{C00000} -1.71} & {\color[HTML]{C00000} -25.12} & {\color[HTML]{C00000} -0.98} & {\color[HTML]{C00000} -5.21} \\ \midrule
Oracle             & 82.32                        & 71.44                        & 71.51                        & 89.19                        & 59.93                        & 86.06                        & 74.62                        & 73.08                        & 89.74                        & 60.8                         & 96.87                        & 68.4                          & 96.94                        & 67.85                        \\
(vs. Ours)         & {\color[HTML]{C00000} -0.02} & {\color[HTML]{C00000} -0.02} & {\color[HTML]{C00000} -0.02} & {\color[HTML]{70AD47} 0.70}  & {\color[HTML]{C00000} -0.25} & {\color[HTML]{70AD47} 1.47}  & {\color[HTML]{70AD47} 0.94}  & {\color[HTML]{70AD47} 0.35}  & {\color[HTML]{70AD47} 0.70}  & {\color[HTML]{C00000} -0.53} & {\color[HTML]{70AD47} 0.41}  & {\color[HTML]{C00000} -0.71}  & {\color[HTML]{70AD47} 0.33}  & {\color[HTML]{C00000} -0.57} \\ \bottomrule
\end{tabular}%
}
\caption{\textit{Ours} are consistently higher than \textit{Random} across all models and benchmarks and are close to \textit{Oracle} (gold accuracy).}
\label{tab:e2e}
\vspace{-1em}
\end{table}

%% file: eval.tex
\section{Evaluation}
\label{sec:eval}
\subsection{Experiment Setup}
\input{fig-budget-acc}
\input{tab-effNonlinear}
\paragraph{Models \& datasets}
For NLP, 
we choose target models as BERT (12 layers) and DistilBERT (6 layers), 
with 12 attention heads and a hidden dimension of 768. 
For vision, 
we chose the target model as ViT, with the same architecture as BERT. 
We chose benchmarks with at least tens of thousands of data points which warrant selection. Following prior work \citep{xu2022data} to construct the training set with imbalanced labels, we remove some data points from the original benchmarks.
Our imbalanced NLP benchmarks include binary classification from GLUE \citep{wang2018glue}: 
SST2 (42K), QNLI (58K), and QQP (149K); 
and multilabel classification~\citep{Zhang2015CharacterlevelCN}: AG news (40K) and Yelp review full (188K).
Our vision benchmarks are two multilabel benchmarks CAFAR-10 (10K) and CIFAR-100 (6K).
%
Note that we do not change the test set. 


We implement our scheme based on Crypten \citep{knott2021crypten},  a popular MPC framework.
We run data/model owners on separate GPUs (Nvidia Quadro RTX 4000) with Intel(R) Xeon(R) Silver 4210 CPU; 
we control the network bandwidth (100 MB/sec) and latency (100 ms) between them to emulate a wide-area network condition. 
We report:
(1) the test accuracy of the target model, after being trained with the selected data; 
(2) the delay of the data selection process over MPC. 

These results are derived using two-phase data selection unless specified otherwise. In the first phase, for NLP tasks, we employ a proxy model with a single layer and a dimensionality of 2, while for CV tasks, a proxy model with three layers and the same dimensionality is utilized. In the second phase, a proxy model with three layers and a dimensionality of 16 is applied for these two tasks.

\paragraph{Baselines} 
We evaluate our method against the following baselines. 
\begin{myenumerate}
    \item \textit{Random} selects data randomly from the data owner. 
    It incurs zero MPC overhead. 
    
    \item \Oracle{} queries data using the target model. 
    Under our framework (\sect{motiv}), 
    we consider that it leads to gold test accuracy. 
    Yet, it incurs prohibitive MPC overhead as will be shown. 
    
    
    \item \textit{\mf} \citep{li2022mpcformer} is a closely related project that minimizes the cost of evaluating transformers over MPC. 
    We implement this baseline to adopt \mf's key optimizations for evaluating \proxy{} models: 
    linear operators for approximating nonlinearity and model distillation for recovering accuracy.
    For a fair comparison, \textit{\mf} uses the same proxy model architecture as ours, and initializes the proxy models in the same way; 
    runs distillation on the same amount of bootstrap data. 
       Note that we cannot directly compare to PUMA \citep{dong2023puma}, which is designed under 3 compute parties. 
    
    \item \textit{Bolt} \citep{pang2024bolt}'s MPC framework contains linear approximations for nonlinear operations.
        
\end{myenumerate}

\input{fig-e2e-latency}
\subsection{End-to-End Results}

\paragraph{Selection delays}
Our method significantly reduces the delays. 
Compared to \oracle{}, 
our method's delay is lower by two orders of magnitude as shown in \autoref{fig:e2e-latency}. 
For instance, to select 20\% from 42K data points (SST2 benchmark, DistilBERT model), 
our experiment takes around 20 hours whereas \textit{Oracle} would take 3740 hours. 


\paragraph{Selection efficacy}
Our method effectively selects training data that best boosts the test accuracy. 
\autoref{tab:e2e} zooms in on a typical purchase budget (20\%). 
Compared to \textit{Random}, 
our test accuracy is consistently higher across all NLP and CV benchmarks. 
The benefit is particularly pronounced on challenging tasks such as CIFAR-100, 
on which our accuracy is higher by 25.11\%/5.21\% on ViT-small/base. 
Compared to \oracle{}, our test accuracy is comparable: 
only 0.08\% -- 0.58\% lower for DistillBERT and BERT, 
and even 0.13\% and 0.15\% higher for ViT-base and ViT-small.

Across different purchase budgets of 25\%, 30\%, and 40\%, our method also leads to strong accuracy. 
Compared to \textit{Random}, 
our accuracy is 2.23\% higher on average, and up to 5.14\%. 
Compared to \oracle{}, 
our average accuracy is only 0.91\% lower on average, and as low as 0.01\%. 
See the Appendix for detailed results. 

From a different angle to view our efficacy: 
to reach a given accuracy, our method demands a much lower budget. 
As shown in \autoref{fig:budget-acc}, to be on par with our test accuracy on a budget of 20\%, 
\textit{Random} would need a budget of 70\% (BERT on QNLI),
a budget of 90\% (DistilBERT on SST2), 
and a budget of 100\% (DistilBERT on YELP and BERT on YELP/SST2).

\input{tab-mpcformer} 
\subsection{Comparison with {\mf} and Bolt}

As shown in \autoref{tab:mpcformer}, 
our method simultaneously delivers much higher test accuracy (by 23\%--36\%) and much shorter (by 7$\times$) selection delays. 
The reasons are twofold. 
(1) \textit{\mf}'s longer delays come from that it approximates individual nonlinear operators with linear counterparts; it does not reduce model dimension as we do. 
(2) \textit{\mf}'s lower accuracy comes from that its \proxy{} models come from the distillation of the target model, a process requiring ample, representative training data. 
This is at odds with data selection: 
$S_{boot}$ is the only data available for distillation; it has a small size and skewed labels. 
The skewness propagates from $S_{boot}$ to \proxy{} models, to selected data, and to the target model. 
As a result, 
the trained target model is biased towards predicting the majority class in the training set. 
This results in poor accuracy if the test and train sets have different majority classes, sometimes even worse than random guess (e.g. QQP in \autoref{tab:mpcformer}). See \autoref{sec:bolt} for the comparison with \textit{Bolt}.

\input{fig-ablation-delay}

\subsection{Ablation Study}

\paragraph{Crypten incurs minor accuracy loss}
We validate that our choice of the Crypten framework \citep{knott2021crypten}, which executes ML on a finite ring, 
introduces minor accuracy loss. 
On SST2 with the default budget (5\% for $S_{boot}$; 20\% total),
private selection fully executed on Crypten results in test accuracy of 83.37\%, 
which is even slightly (0.5\%) higher than executing on PyTorch.

\paragraph{MLP for non-linearity}
(1) Delay. Our use of MLP reduces the selection delay by two orders of magnitude, 
e.g. from 3740 hours to around 20 hours for DistilBERT on SST2. 
This is shown in Figure \ref{fig:ablation-delay}, by the difference between \textit{P} and \textit{PM}. 
(2) Accuracy. The use of MLP incurs minor degradation in the test accuracy. 
The results are shown in \autoref{tab:mlp}, by the difference between \textit{Ours} to \textit{NoApprox}. 
On SST2, ours is 0.78\% higher with DistilBERT and only 0.46\% lower with BERT. On AGNEWS, ours are just 1.24\% lower with DistilBERT and 0.93\% lower with BERT. On QQP, it's 1.21\% and 0.05\% higher respectively.
Among the three MLPs, softmax in the attention module shows a higher impact on accuracy than the other two. 
As shown in \autoref{tab:mlp} and across all the datasets, by the difference between \textit{Ours} to \textit{NoAttnSM}.
Using softmax approximation MLP is just 0.75\% lower with DistilBERT and 0.99\% lower with BERT on average;
by the difference between \textit{Ours} to \textit{NoAttnLN}, our approximations of LayerNorm only decreases the performance by 0.76\% with DistillBERT and 0.16\% higher with BERT on average.
This is justified by the higher cost that our AttentionSM reduces data communication by 42$\times$, while AttentionLN only reduces 8.25$\times$.




\paragraph{Efficacy of multi-phase selection (MPS)}
\input{tab-multiphase-acc}
We compare MPS to single-phase selection (SPS), 
in which the \proxy{} model is the same as the one used in the last phase of MPS; 
it has all our other optimizations.  
(1) \Mps{} reduces delays significantly. 
Compared to \sps{}, the two-phase selection
reduces the total delay by 33\%--61\% across all the benchmarks. 
Because \mps{} runs a much smaller proxy model in phase 1 to filter most datapoints. 
The advantage is more pronounced for large models such as BERT, 
where the efficiency gap between the proxy models in phase 1 is even larger. 
(2) \Mps{} improves accuracy moderately. 
We focus on two representative benchmarks SST2 and QQP (binary and multi-label classification).
The results are in \autoref{tab:multiphase-acc}.
Going from one phase to two/three phases results in tangible accuracy gain (around 1\%); 
notably BERT sees accuracy gain as high as 1.72\% (SST2) and 8.59\% (QQP). 
We attribute the gain to \mps{} being more selective for the finally selected datapoints, which have been sieved through multiple proxy models, small and big. 

\paragraph{Sizes of proxy models}
Multi-layer proxy models increase the selection delay while may improve the test accuracy. 
Our experiments show that: 
for small target models on challenging benchmarks (e.g. ViT-small on CIFAR-100), 
the test accuracy from a three-layer proxy outperforms a one-layer proxy significantly 
(by 17.33\%); 
on other benchmarks, 1+ layers only result in a minor increase in the test accuracy. 
On CIFAR-100, a three-layer ViT-base proxy model is 11.32\% higher than a one-layer with 66.6\% lower efficiency. Their accuracies are 0.26\% higher than Oracle's on average, but with 4$\times$ higher efficiency. It is worth trading some extent of efficiency for much higher accuracy by a multi-layer proxy model, especially on difficult datasets like CIFAR-100.




\paragraph{MLP hidden dimensions}
To decide $d_i$, we conduct a grid search to find the best values, which are included in our selection schedules shown in \autoref{tab:multiphase-acc}. We mainly do this grid search on SST2, validate the hyperparameters on QQP before applying the same settings to other datasets.
A higher dimension like 16 has good accuracy on different datasets while not incurring much more latency. Multi-phase selection can reduce lots of latency, hence, we pick hidden dimension combinations that can bring more accuracy gain. For the 2 Phase, we choose (2, 16) which has the highest accuracy on larger datasets and target models, such as BERT on QQP. For the 3 Phase, we choose (2, 8, 16). It has the highest accuracy with DisitilBERT on SST2, with BERT on QQP, and suboptimal accuracy on other experiments.

\paragraph{IO scheduling}
This optimization reduces the end-to-end selection delays by 1.3-1.4$\times$, 
as shown in Figure \ref{fig:ablation-delay}, the difference between \textit{Ours} and \textit{PMT}. 
The reduction comes from overlapping the MPC computation 
and the communication rounds. 
their communication latency is hidden. 
The reduction depends on hardware: 
device computation bandwidth and the network latencies. 

%% file: fig-budget-acc.tex

\begin{figure*}[t]
	\centering
	\includegraphics[width=0.9\textwidth]{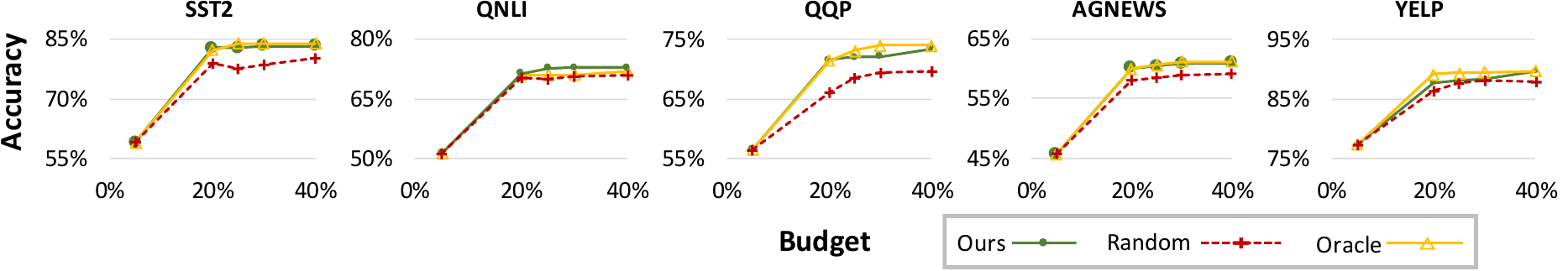}
	\caption{Across a variety of budgets, \textit{Ours} consistently outperforms \textit{Random} and are comparable with \textit{Oracle} (gold accuracy). Target model: DistilBERT.}
	\label{fig:budget-acc}
\end{figure*}

%% file: tab-effNonlinear.tex
\begin{table*}[t]
	\centering
	\includegraphics[width=\textwidth{}]{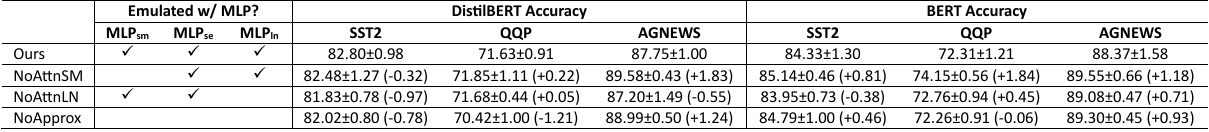}	
 \vspace{-1.5em}
	\caption{Using MLPs to emulate Transformer nonlinearity results in minor accuracy degradation 
	}
	\label{tab:effNonlinear}
	\label{tab:mlp}
\end{table*}

%% file: fig-e2e-latency.tex

\begin{figure*}[t]
	
  \centering
	\includegraphics[width=0.7\textwidth{}]{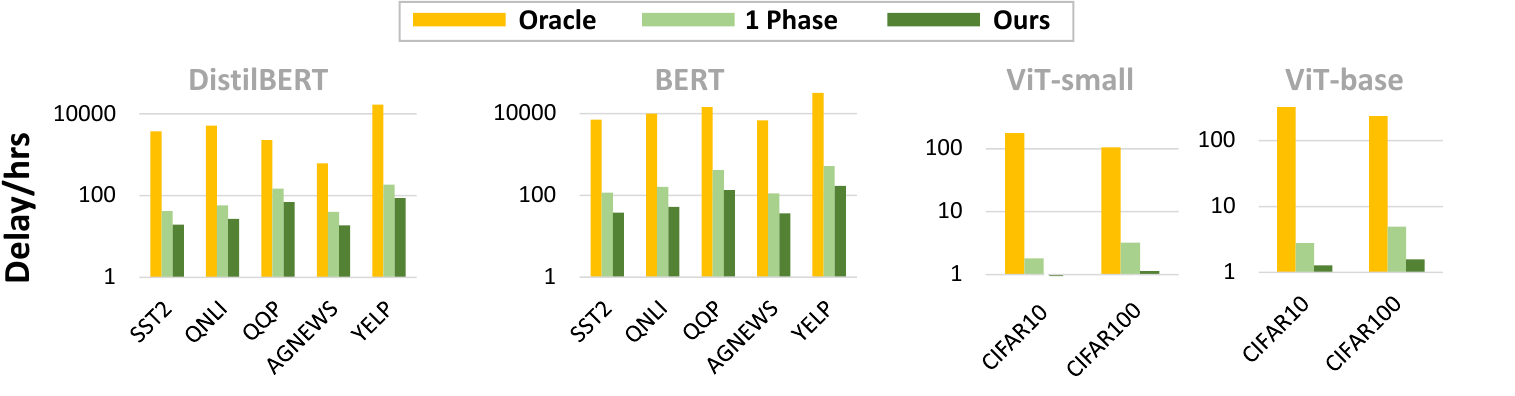}
  \caption{
  Our method reduces the end-to-end delays by two orders of magnitude. As in Section \ref{sec:detail}, 1 phase selection selects data using a proxy model with a hidden dimension of 16.
}
  \label{fig:e2e-latency}
  \vspace{-1em}
\end{figure*}


%% file: tab-mpcformer.tex
\begin{wraptable}{r}{0.5\textwidth}
\centering
\includegraphics[width=0.5\textwidth{}]{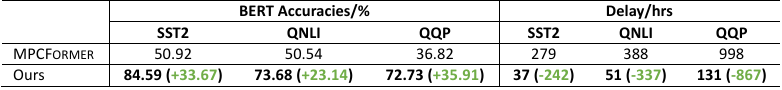}
    \vspace{-1em}
    \caption{Compared to \mf, our method results in much higher test accuracy and much lower delays.    
    On GLUE benchmarks which were also evaluated and reported by \mf.
    Target model: BERT. Proxy model: 1 transformer layer with 1 head + 3 transformer layers with 12 heads.}
    \label{tab:mpcformer}
    \vspace{-1.5em}
\end{wraptable} 

%% file: fig-ablation-delay.tex
 \begin{figure*}[t]
 	 \centering
 	 \includegraphics[width=0.7\textwidth]{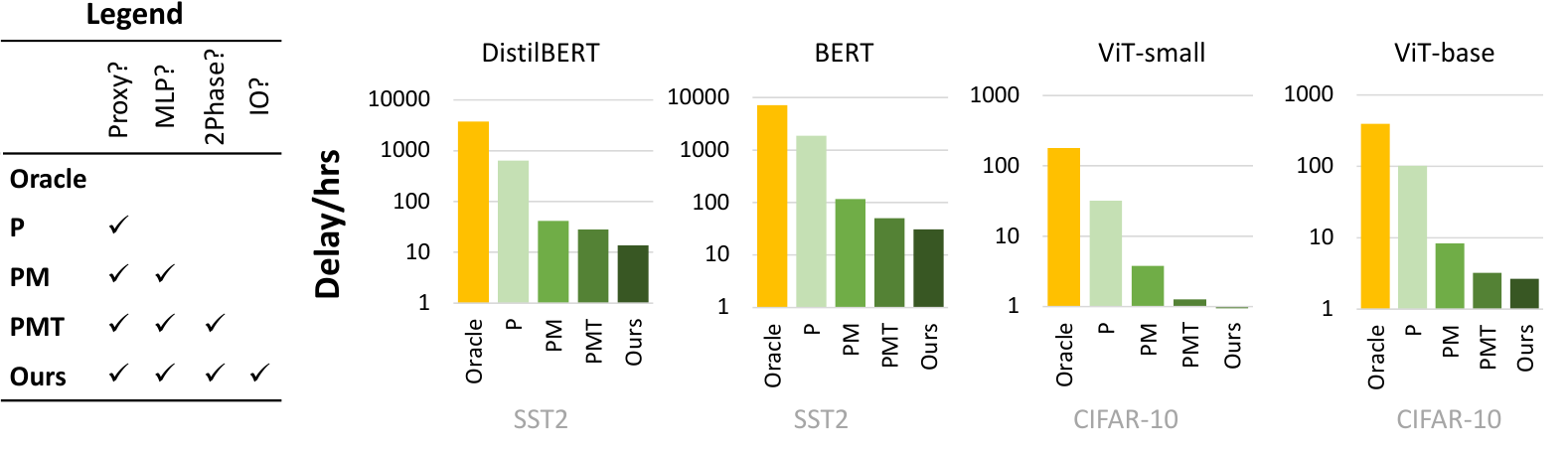}
 	 \caption{Delay reduction by our techniques.
   The results also show that using proxy models alone (variant ``P'' in the figure,) is not enough.}
 	 \label{fig:ablation-delay}
   \vspace{-1em}
 \end{figure*}







%% file: tab-multiphase-acc.tex
\begin{table*}[t]
	\centering
	\includegraphics[width=\textwidth{}]{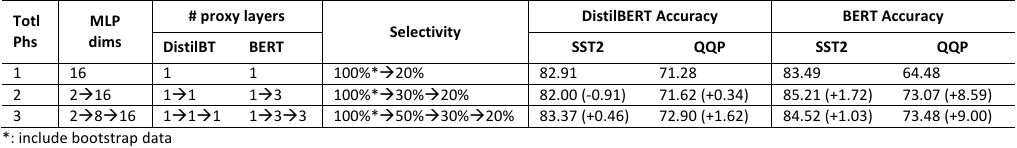}
	\vspace{-1em}
	\caption{Our multi-phase selection improves accuracy in general. 
	Three example schedules (with 1, 2, and 3 total phases, respectively) and their test accuracies are shown.}
	\label{tab:multiphase-acc}
	\vspace{-1em}
\end{table*}

%% file: summary.tex
\section{Concluding remarks}

This paper addresses the private selection of imbalanced, unlabelled data. We propose data selection on MPC, preserving privacy while evaluating data. Overcoming computational challenges, we introduce a data-aware MLP approximation for non-linear modules, tailored to different datasets for reduced latency. Our multi-phase selection mechanism enhances proxy models' efficiency across diverse benchmarks.


\section*{Acknowledgment}
The authors were supported in part by NSF awards \#2128725, \#1919197, and \#2106893; 
as well as NSF awards \#2124538 and \#2007492.
The authors thank the anonymous reviewers for their
insightful feedback.

%% file: appendix.tex
\section{Appendix}
\label{sec:appendix}
\subsection{MLP Approximation is tied to data selection}
Inspired by the outstanding performance of our MLP approximations for nonlinear modules, one question was raised: is the good performance of MLP approximations because of its inference capability? Our further experiments show that our MLP approximation is tied to data selection, instead of general model inference.

Following \citet{li2022mpcformer}, we replace three nonlinear modules in a BERT-base model with our approximations: Attention Softmax, Attention LayerNorm, and Feedforward Network Layernorm. There are 3*12 = 36 MLPs in the BERT model. In the experiments, two LayerNorm approximations are always data-aware, while there are both data-aware and fixed approximations of Softmax because of varying attention masks. The BERT with MLP approximation will be fine-tuned on the benchmark before inference.

Three versions of attention softmax approximations achieve no better than random guess performances: (1) Using fixed MLP for attention softmax and changing the mask value to -3 instead of a very negative value for simpler distribution. Our BERT with approximation achieves only 52.92\% accuracy. (2) Using data-aware attention softmax approximation, which zeros out masked values on MLP outputs, has 49.08\% accuracy. If we normalize the remaining values to [0, 1], the accuracy remains 49.08\%. To better understand the influence of each nonlinear approximation, we did an ablation study that keeps just one kind of MLPs (12 of them) in BERT. However, none of them achieve better than 50.92\% accuracies. We further notice that adding just one attention softmax MLP to the model has no impact on the inference accuracy. But adding one layernorm MLP will degrade the accuracy by 0.85\% on average. These results show that having just one MLP will hurt the inference performance a lot; having 12 or even 36 MLPs will degrade model performance drastically. (3) Removing the attention mask to make MLP data-aware. It always has 50.92\% accuracies, no matter with all three approximations or just one approximation.

The poor inference performance and good data selection performance show that the MLP approximation is specifically suitable for data selection while impractical for model inference directly.

\subsection{Comparison with Bolt}
\label{sec:bolt}
We compare with \textit{Bolt} \citep{pang2024bolt}'s polynomial softmax approximation by replacing MLPs in our 2-phase proxy models while keeping other nonlinear layers intact. This approach ensures the highest softmax accuracy but the largest delay. \textit{Bolt} achieves an accuracy of $69.04\%\pm11.86\%$, ours achieves $84.59\%\pm0.95\%$, and \textit{\mf} achieves 50.92\%. With proxy models trained on $S_{boot}$, \textit{Bolt} outperforms \textit{\mf}. However, our method surpasses \textit{Bolt} by 15.55\% in accuracy and had a lower standard deviation, demonstrating better robustness against the biased $S_{boot}$. Notably, if full approximations (e.g., reciprocal and layer normalization) are applied, \textit{Bolt}'s accuracy is expected to decrease further, reducing its effectiveness compared to our methods.

\subsection{Additional Experiments Results}

\input{fig-tradeoff}
\input{tab-gridsearch}
\input{tab-budget}
\input{tab-randomselect}
\begin{table*}[ht]
	\centering
	\includegraphics[width=\textwidth{}]{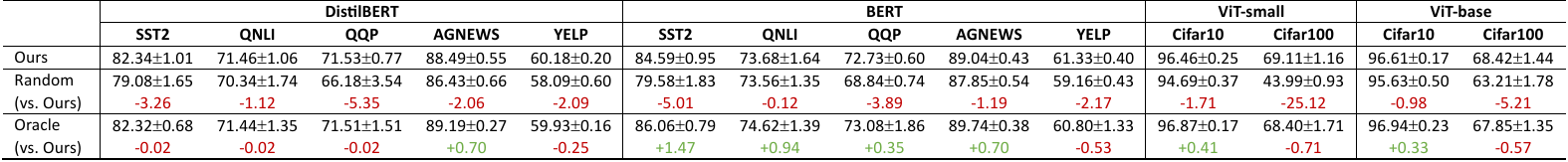}
	\vspace{-1em}
	\caption{\textit{Ours} are consistently higher than \textit{Random} across all models and benchmarks and are close to \textit{Oracle} (gold accuracy). Compared with \autoref{tab:e2e}, standard deviations of 5 runs are added.}
	\label{tab:appendixe2e}
\end{table*}

%% file: fig-tradeoff.tex
\begin{figure}[ht]
	\centering
	\includegraphics[width=\textwidth]{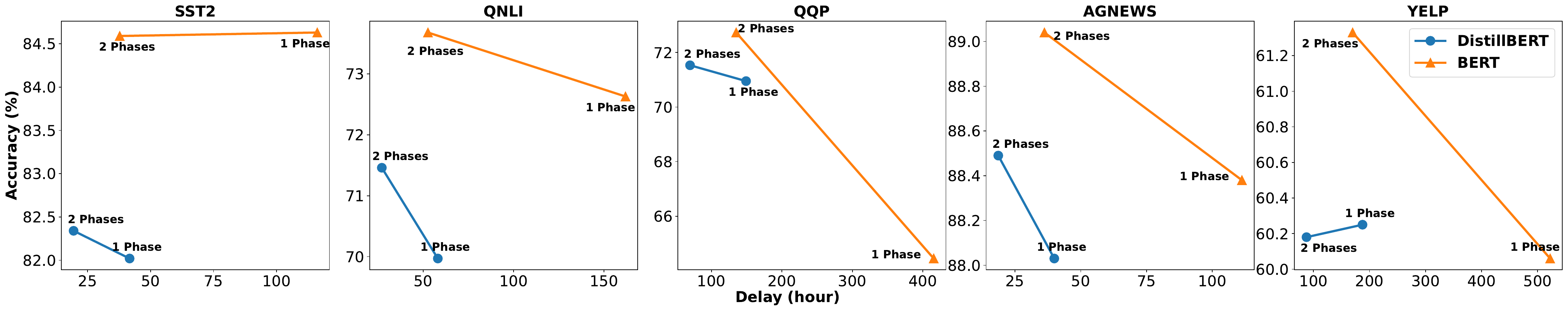}
		\vspace{-2em}
	\caption{Multi-phase selection achieves a better trade-off between accuracy and delay in data selection. 2-phase proxy model: 1 layer with 1 attention head and a dimension of 2, 3 layers with 12 attention heads and a dimension of 16; 1-phase proxy model: 3 layers with 12 attention heads and a dimension of 16.}
	\label{fig:tradeoff}
\end{figure}

%% file: tab-gridsearch.tex
\begin{table}[ht]
	\centering
	\includegraphics[width=0.5\textwidth{}]{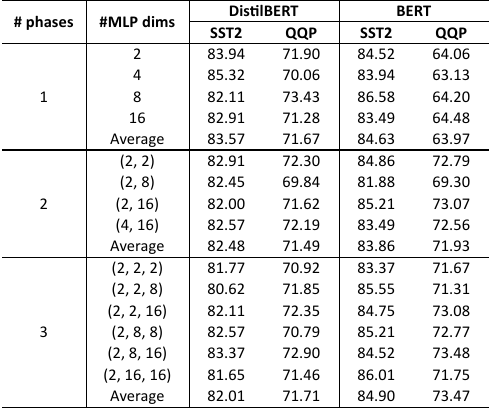}
	\caption{Multi-phase selection is able to significantly reduce runtime and maintain comparable accuracy. On QQP, three-phase selection achieves 9.5\% higher accuracy than using one phase.
	Main results: 16, (2,16), (2,8,16).
	}
	\label{tab:gridsearch}
\end{table}

%% file: tab-budget.tex
\begin{table*}[h]
	\includegraphics[width=\textwidth{}]{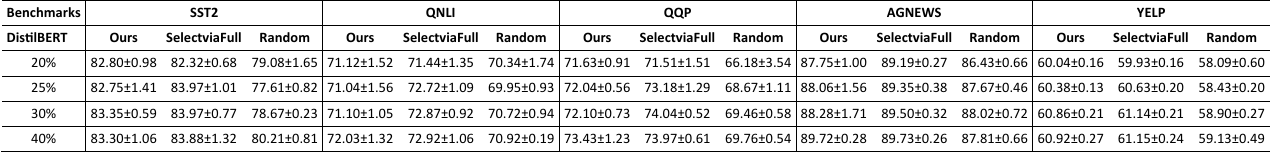}
	\caption{
 Our methods are robust against the increase in purchase budget. We have consistently better accuracy than Random by a large margin and achieve comparable performance with Oracle, up to 40\% budget. Ours: 2-phase selection. 1 layer with 1 attention head and a dimension of 2 in phase 1, 3 layers with 12 attention heads and a dimension of 16 in phase 2.
 }
	\label{tab:budget}
\end{table*}

%% file: tab-randomselect.tex
\begin{table*}[h]
	\includegraphics[width=1.0\textwidth{}]{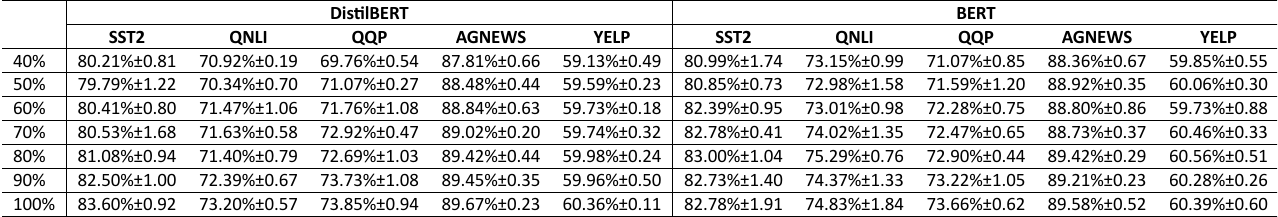}
	\caption{Accuracies of randomly selecting more data to train the target model. Compared with our selection accuracies in the main body, Random needs much more data. 100\% YELP and 90\% QQP data is necessary for our 20\% selection with DistilBERT; 100\% YELP and SST2, 80\% QQP and AGNEWS are necessary to match our 20\% budget performance of BERT.}
	\label{tab:randomselect}
\end{table*}

%% file: main.bbl
\begin{thebibliography}{34}
\providecommand{\natexlab}[1]{#1}
\providecommand{\url}[1]{\texttt{#1}}
\expandafter\ifx\csname urlstyle\endcsname\relax
  \providecommand{\doi}[1]{doi: #1}\else
  \providecommand{\doi}{doi: \begingroup \urlstyle{rm}\Url}\fi

\bibitem[Agrawal et~al.(2019)Agrawal, Shahin~Shamsabadi, Kusner, and
  Gasc{\'o}n]{agrawal2019quotient}
Nitin Agrawal, Ali Shahin~Shamsabadi, Matt~J Kusner, and Adri{\`a} Gasc{\'o}n.
\newblock Quotient: two-party secure neural network training and prediction.
\newblock In \emph{Proceedings of the 2019 ACM SIGSAC Conference on Computer
  and Communications Security}, pp.\  1231--1247, 2019.

\bibitem[Beaver(1992)]{Beavor1992}
Donald Beaver.
\newblock Efficient multiparty protocols using circuit randomization.
\newblock In Joan Feigenbaum (ed.), \emph{Advances in Cryptology --- CRYPTO
  '91}, pp.\  420--432, Berlin, Heidelberg, 1992. Springer Berlin Heidelberg.
\newblock ISBN 978-3-540-46766-3.

\bibitem[Chen et~al.(2021)Chen, Zhang, Ouyang, Liu, Shen, and
  Wang]{Chen_2021_CVPR}
Tianlong Chen, Zhenyu Zhang, Xu~Ouyang, Zechun Liu, Zhiqiang Shen, and
  Zhangyang Wang.
\newblock "bnn - bn = ?": Training binary neural networks without batch
  normalization.
\newblock In \emph{Proceedings of the IEEE/CVF Conference on Computer Vision
  and Pattern Recognition (CVPR) Workshops}, pp.\  4619--4629, June 2021.

\bibitem[Chen et~al.(2022)Chen, Bao, Huang, Dong, Jiao, Jiang, Zhou, Li, and
  Wei]{chen-etal-2022-x}
Tianyu Chen, Hangbo Bao, Shaohan Huang, Li~Dong, Binxing Jiao, Daxin Jiang,
  Haoyi Zhou, Jianxin Li, and Furu Wei.
\newblock {THE}-{X}: Privacy-preserving transformer inference with homomorphic
  encryption.
\newblock In \emph{Findings of the Association for Computational Linguistics:
  ACL 2022}, pp.\  3510--3520, Dublin, Ireland, May 2022. Association for
  Computational Linguistics.
\newblock \doi{10.18653/v1/2022.findings-acl.277}.
\newblock URL \url{https://aclanthology.org/2022.findings-acl.277}.

\bibitem[Chou et~al.(2018)Chou, Beal, Levy, Yeung, Haque, and
  Fei-Fei]{chou2018faster}
Edward Chou, Josh Beal, Daniel Levy, Serena Yeung, Albert Haque, and
  Li~Fei-Fei.
\newblock Faster cryptonets: Leveraging sparsity for real-world encrypted
  inference.
\newblock \emph{arXiv preprint arXiv:1811.09953}, 2018.

\bibitem[Coleman et~al.(2019)Coleman, Yeh, Mussmann, Mirzasoleiman, Bailis,
  Liang, Leskovec, and Zaharia]{coleman2019selection}
Cody Coleman, Christopher Yeh, Stephen Mussmann, Baharan Mirzasoleiman, Peter
  Bailis, Percy Liang, Jure Leskovec, and Matei Zaharia.
\newblock Selection via proxy: Efficient data selection for deep learning.
\newblock \emph{arXiv preprint arXiv:1906.11829}, 2019.

\bibitem[Dong et~al.(2023)Dong, Lu, Zheng, Wu, Zhao, Tan, Huang, Hong, Wei, and
  Cheng]{dong2023puma}
Ye~Dong, Wen-jie Lu, Yancheng Zheng, Haoqi Wu, Derun Zhao, Jin Tan, Zhicong
  Huang, Cheng Hong, Tao Wei, and Wenguang Cheng.
\newblock Puma: Secure inference of llama-7b in five minutes.
\newblock \emph{arXiv preprint arXiv:2307.12533}, 2023.

\bibitem[Fu et~al.(2021)Fu, Yu, Zhang, Wu, Ouyang, Cox, and
  Lin]{NEURIPS2021_6ce8d8f3}
Yonggan Fu, Qixuan Yu, Yang Zhang, Shang Wu, Xu~Ouyang, David Cox, and Yingyan
  Lin.
\newblock Drawing robust scratch tickets: Subnetworks with inborn robustness
  are found within randomly initialized networks.
\newblock In M.~Ranzato, A.~Beygelzimer, Y.~Dauphin, P.S. Liang, and J.~Wortman
  Vaughan (eds.), \emph{Advances in Neural Information Processing Systems},
  volume~34, pp.\  13059--13072. Curran Associates, Inc., 2021.
\newblock URL
  \url{https://proceedings.neurips.cc/paper_files/paper/2021/file/6ce8d8f3b038f737cefcdafcf3752452-Paper.pdf}.

\bibitem[Fu et~al.(2022)Fu, Yu, Li, Ouyang, Chandra, and
  Lin]{fu2022contrastive}
Yonggan Fu, Qixuan Yu, Meng Li, Xu~Ouyang, Vikas Chandra, and Yingyan Lin.
\newblock Contrastive quant: quantization makes stronger contrastive learning.
\newblock In \emph{Proceedings of the 59th ACM/IEEE Design Automation
  Conference}, pp.\  205--210, 2022.

\bibitem[Goldreich et~al.(1987)Goldreich, Micali, and
  Wigderson]{Goldreich1987secretshare}
O.~Goldreich, S.~Micali, and A.~Wigderson.
\newblock How to play any mental game.
\newblock In \emph{Proceedings of the Nineteenth Annual ACM Symposium on Theory
  of Computing}, STOC '87, pp.\  218–229, New York, NY, USA, 1987.
  Association for Computing Machinery.
\newblock ISBN 0897912217.
\newblock \doi{10.1145/28395.28420}.
\newblock URL \url{https://doi.org/10.1145/28395.28420}.

\bibitem[Goodfellow et~al.(2016)Goodfellow, Bengio, and
  Courville]{Goodfellow-et-al-2016}
Ian Goodfellow, Yoshua Bengio, and Aaron Courville.
\newblock \emph{Deep Learning}.
\newblock MIT Press, 2016.
\newblock \url{http://www.deeplearningbook.org}.

\bibitem[Hoi et~al.(2006)Hoi, Jin, Zhu, and Lyu]{Hoi2006MedicalImage}
Steven C.~H. Hoi, Rong Jin, Jianke Zhu, and Michael~R. Lyu.
\newblock Batch mode active learning and its application to medical image
  classification.
\newblock In \emph{Proceedings of the 23rd International Conference on Machine
  Learning}, ICML '06, pp.\  417–424, New York, NY, USA, 2006. Association
  for Computing Machinery.
\newblock ISBN 1595933832.
\newblock \doi{10.1145/1143844.1143897}.
\newblock URL \url{https://doi.org/10.1145/1143844.1143897}.

\bibitem[Hornik et~al.(1989)Hornik, Stinchcombe, and
  White]{hornik1989multilayer}
Kurt Hornik, Maxwell Stinchcombe, and Halbert White.
\newblock Multilayer feedforward networks are universal approximators.
\newblock \emph{Neural networks}, 2\penalty0 (5):\penalty0 359--366, 1989.

\bibitem[Katharopoulos \& Fleuret(2018)Katharopoulos and
  Fleuret]{katharopoulos2018not}
Angelos Katharopoulos and Fran{\c{c}}ois Fleuret.
\newblock Not all samples are created equal: Deep learning with importance
  sampling.
\newblock In \emph{International conference on machine learning}, pp.\
  2525--2534. PMLR, 2018.

\bibitem[Kaur et~al.(2019)Kaur, Pannu, and Malhi]{kaur2019systematic}
Harsurinder Kaur, Husanbir~Singh Pannu, and Avleen~Kaur Malhi.
\newblock A systematic review on imbalanced data challenges in machine
  learning: Applications and solutions.
\newblock \emph{ACM Computing Surveys (CSUR)}, 52\penalty0 (4):\penalty0 1--36,
  2019.

\bibitem[Knott et~al.(2021)Knott, Venkataraman, Hannun, Sengupta, Ibrahim, and
  van~der Maaten]{knott2021crypten}
Brian Knott, Shobha Venkataraman, Awni Hannun, Shubho Sengupta, Mark Ibrahim,
  and Laurens van~der Maaten.
\newblock Crypten: Secure multi-party computation meets machine learning.
\newblock \emph{Advances in Neural Information Processing Systems},
  34:\penalty0 4961--4973, 2021.

\bibitem[Li et~al.(2022)Li, Shao, Wang, Guo, Xing, and Zhang]{li2022mpcformer}
Dacheng Li, Rulin Shao, Hongyi Wang, Han Guo, Eric~P Xing, and Hao Zhang.
\newblock Mpcformer: fast, performant and private transformer inference with
  mpc.
\newblock \emph{arXiv preprint arXiv:2211.01452}, 2022.

\bibitem[Mahmood et~al.(2022)Mahmood, Lucas, Alvarez, Fidler, and
  Law]{Mahmood2022OptimizingDataCollection}
Rafid Mahmood, James Lucas, Jose~M. Alvarez, Sanja Fidler, and Marc Law.
\newblock Optimizing data collection for machine learning.
\newblock In S.~Koyejo, S.~Mohamed, A.~Agarwal, D.~Belgrave, K.~Cho, and A.~Oh
  (eds.), \emph{Advances in Neural Information Processing Systems}, volume~35,
  pp.\  29915--29928. Curran Associates, Inc., 2022.
\newblock URL
  \url{https://proceedings.neurips.cc/paper_files/paper/2022/file/c1449acc2e64050d79c2830964f8515f-Paper-Conference.pdf}.

\bibitem[Mindermann et~al.(2022)Mindermann, Brauner, Razzak, Sharma, Kirsch,
  Xu, H{\"o}ltgen, Gomez, Morisot, Farquhar, et~al.]{mindermann2022prioritized}
S{\"o}ren Mindermann, Jan~M Brauner, Muhammed~T Razzak, Mrinank Sharma, Andreas
  Kirsch, Winnie Xu, Benedikt H{\"o}ltgen, Aidan~N Gomez, Adrien Morisot,
  Sebastian Farquhar, et~al.
\newblock Prioritized training on points that are learnable, worth learning,
  and not yet learnt.
\newblock In \emph{International Conference on Machine Learning}, pp.\
  15630--15649. PMLR, 2022.

\bibitem[Mirzasoleiman et~al.(2020)Mirzasoleiman, Bilmes, and
  Leskovec]{mirzasoleiman2020coresets}
Baharan Mirzasoleiman, Jeff Bilmes, and Jure Leskovec.
\newblock Coresets for data-efficient training of machine learning models.
\newblock In \emph{International Conference on Machine Learning}, pp.\
  6950--6960. PMLR, 2020.

\bibitem[Mishra et~al.(2020)Mishra, Lehmkuhl, Srinivasan, Zheng, and
  Popa]{244032}
Pratyush Mishra, Ryan Lehmkuhl, Akshayaram Srinivasan, Wenting Zheng, and
  Raluca~Ada Popa.
\newblock Delphi: A cryptographic inference service for neural networks.
\newblock In \emph{29th USENIX Security Symposium (USENIX Security 20)}, pp.\
  2505--2522. USENIX Association, August 2020.
\newblock ISBN 978-1-939133-17-5.
\newblock URL
  \url{https://www.usenix.org/conference/usenixsecurity20/presentation/mishra}.

\bibitem[Mohassel \& Zhang(2017)Mohassel and Zhang]{mohassel2017secureml}
Payman Mohassel and Yupeng Zhang.
\newblock Secureml: A system for scalable privacy-preserving machine learning.
\newblock In \emph{2017 IEEE symposium on security and privacy (SP)}, pp.\
  19--38. IEEE, 2017.

\bibitem[Ouyang et~al.(2022)Ouyang, Ansari, Lin, and Ji]{ouyang2022efficient}
Xu~Ouyang, Shahina Mohd~Azam Ansari, Felix~Xiaozhu Lin, and Yangfeng Ji.
\newblock Efficient nlp model finetuning via multistage data filtering.
\newblock \emph{arXiv preprint arXiv:2207.14386}, 2022.

\bibitem[Pang et~al.(2024)Pang, Zhu, M{\"o}llering, Zheng, and
  Schneider]{pang2024bolt}
Qi~Pang, Jinhao Zhu, Helen M{\"o}llering, Wenting Zheng, and Thomas Schneider.
\newblock Bolt: Privacy-preserving, accurate and efficient inference for
  transformers.
\newblock In \emph{2024 IEEE Symposium on Security and Privacy (SP)}, pp.\
  4753--4771. IEEE, 2024.

\bibitem[Settles(2012)]{Settles2012Activelearning}
Burr Settles.
\newblock \emph{Active Learning}.
\newblock Synthesis Lectures on Artificial Intelligence and Machine Learning.
  Morgan \& Claypool Publishers, 2012.

\bibitem[Shamir(1979)]{shamir1979share}
Adi Shamir.
\newblock How to share a secret.
\newblock \emph{Communications of the ACM}, 22\penalty0 (11):\penalty0
  612--613, 1979.

\bibitem[Smith et~al.(2018)Smith, Nebgen, Lubbers, Isayev, and
  Roitberg]{Smith2018LessIsMore}
Justin~S. Smith, Ben Nebgen, Nicholas Lubbers, Olexandr Isayev, and Adrian~E.
  Roitberg.
\newblock {Less is more: Sampling chemical space with active learning}.
\newblock \emph{The Journal of Chemical Physics}, 148\penalty0 (24):\penalty0
  241733, 05 2018.
\newblock ISSN 0021-9606.
\newblock \doi{10.1063/1.5023802}.
\newblock URL \url{https://doi.org/10.1063/1.5023802}.

\bibitem[Vaswani et~al.(2017)Vaswani, Shazeer, Parmar, Uszkoreit, Jones, Gomez,
  Kaiser, and Polosukhin]{Ashish2017transformer}
Ashish Vaswani, Noam Shazeer, Niki Parmar, Jakob Uszkoreit, Llion Jones,
  Aidan~N. Gomez, Lukasz Kaiser, and Illia Polosukhin.
\newblock Attention is all you need.
\newblock \emph{CoRR}, abs/1706.03762, 2017.
\newblock URL \url{http://arxiv.org/abs/1706.03762}.

\bibitem[Wang et~al.(2018)Wang, Singh, Michael, Hill, Levy, and
  Bowman]{wang2018glue}
Alex Wang, Amanpreet Singh, Julian Michael, Felix Hill, Omer Levy, and Samuel~R
  Bowman.
\newblock Glue: A multi-task benchmark and analysis platform for natural
  language understanding.
\newblock \emph{arXiv preprint arXiv:1804.07461}, 2018.

\bibitem[Xu et~al.(2022)Xu, Hannun, and Van Der~Maaten]{xu2022data}
Xinlei Xu, Awni Hannun, and Laurens Van Der~Maaten.
\newblock Data appraisal without data sharing.
\newblock In \emph{International Conference on Artificial Intelligence and
  Statistics}, pp.\  11422--11437. PMLR, 2022.

\bibitem[Yang et~al.(2019)Yang, Liu, Chen, and Tong]{Qiang20192PC}
Qiang Yang, Yang Liu, Tianjian Chen, and Yongxin Tong.
\newblock Federated machine learning: Concept and applications.
\newblock \emph{CoRR}, abs/1902.04885, 2019.
\newblock URL \url{http://arxiv.org/abs/1902.04885}.

\bibitem[Yao(1986)]{yao1986generate}
Andrew Chi-Chih Yao.
\newblock How to generate and exchange secrets.
\newblock In \emph{27th annual symposium on foundations of computer science
  (Sfcs 1986)}, pp.\  162--167. IEEE, 1986.

\bibitem[You et~al.(2022)You, Li, Sun, Ouyang, and Lin]{you2022supertickets}
Haoran You, Baopu Li, Zhanyi Sun, Xu~Ouyang, and Yingyan Lin.
\newblock Supertickets: Drawing task-agnostic lottery tickets from supernets
  via jointly architecture searching and parameter pruning.
\newblock In \emph{European Conference on Computer Vision}, pp.\  674--690.
  Springer, 2022.

\bibitem[Zhang et~al.(2015)Zhang, Zhao, and LeCun]{Zhang2015CharacterlevelCN}
Xiang Zhang, Junbo~Jake Zhao, and Yann LeCun.
\newblock Character-level convolutional networks for text classification.
\newblock In \emph{NIPS}, 2015.

\end{thebibliography}
